\documentclass[10pt,twocolumn,letterpaper]{article}

\usepackage{cvpr}              
\usepackage{lineno}
\usepackage{nicefrac}
\usepackage{algorithm}
\usepackage{algpseudocodex}

\usepackage{url}            
\usepackage{booktabs}       
\usepackage{amsfonts}       
\usepackage{nicefrac}       
\usepackage{microtype}      
\usepackage{xcolor}         

\usepackage{graphicx} 
\usepackage{tabularx}

\usepackage{multirow}
\usepackage[accsupp]{axessibility}

\usepackage{wrapfig}

\usepackage{colortbl}
\usepackage{url}

\usepackage{xspace}

\definecolor{cvprblue}{rgb}{0.21,0.49,0.74}
\usepackage[pagebackref,breaklinks,colorlinks,allcolors=cvprblue]{hyperref}

\def\paperID{5020} 
\def\confName{CVPR}
\def\confYear{2025}

\usepackage[dvipsnames]{xcolor}

\usepackage{acronym}
\acrodef{ai}[AI]{Artificial Intelligence}
\acrodef{cv}[CV]{Computer Vision}
\acrodef{nlp}[NLP]{Natural Language Processing}
\acrodef{genai}[GenAI]{Generative AI}
\acrodef{llm}[LLM]{Large Language Model}
\acrodef{llms}[LLMs]{Large Language Models}
\acrodef{mllm}[M-LLM]{Multi-Modal Large Language Model}
\acrodef{mllms}[M-LLMs]{Multi-Modal Large Language Models}
\acrodef{videollm}[video-LLM]{video Large Language Model}
\acrodef{lora}[LoRA]{Low Rank Adaptation}
\acrodef{vlm}[VLM]{Vision-Language Models}
\acrodef{vqa}[VQA]{Visual Question Answering}
\acrodef{mlp}[MLP]{Multi Layer Perceptron}
\acrodef{fps}[FPS]{frames per second}

\usepackage{xspace}
\makeatletter
\DeclareRobustCommand\onedot{\futurelet\@let@token\@onedot}
\def\@onedot{\ifx\@let@token.\else.\null\fi\xspace}

\title{M-LLM Based Video Frame Selection for Efficient Video Understanding}

\author{
Kai Hu$^{1,}$\thanks{This work is done during this author's internship at Amazon.} \:\: 
Feng Gao$^{3}$ \:\: 
Xiaohan Nie$^{3}$ \:\: 
Peng Zhou$^{3}$ \:\: 
Son Tran$^{3}$ \:\: 
Tal Neiman$^{3}$  \\ 
Lingyun Wang$^{3}$ \:\: 
Mubarak Shah$^{2,3}$ \:\: 
Raffay Hamid$^{3}$ \:\: 
Bing Yin$^{3}$ \:\: 
Trishul Chilimbi$^{3}$ \\
Carnegie Mellon University $^1$ \qquad
University of Central Florida $^2$  \qquad
Amazon $^3$ \\
{\tt\small kaihu@cs.cmu.edu$^{1,\ast}$} \tt\small{shah@crcv.ucf.edu$^{2}$} \\
{\tt\small{fenggo,nxiaohan,zhoupz,sontran,taneiman,lingyunw,raffay,alexbyin,trishulc\}@amazon.com$^{3}$}}}

\begin{document}
\maketitle

\begin{abstract}
Recent advances in \acf{mllms} show promising results in video reasoning. Popular \ac{mllm} frameworks usually apply naive uniform sampling to reduce the number of video frames that are fed into an \ac{mllm}, particularly for long context videos. However, it could lose crucial context in certain periods of a video, so that the downstream \ac{mllm} may not have sufficient visual information to answer a question. To attack this pain point, we propose a light-weight \ac{mllm}-based frame selection method that adaptively select frames that are more relevant to users' queries. In order to train the proposed frame selector, we introduce two supervision signals (i) Spatial signal, where single frame importance score by prompting an \ac{mllm}; (ii) Temporal signal, in which multiple frames selection by prompting \ac{llm} using the captions of all frame candidates. The selected frames are then digested by a frozen downstream video \ac{mllm} for visual reasoning and question answering. Empirical results show that the proposed \ac{mllm} video frame selector improves the performances various downstream \ac{videollm} across medium (ActivityNet, NExT-QA) and long (EgoSchema, LongVideoBench) context video question answering benchmarks.
\end{abstract} 
\vspace{-9pt}
\section{Introduction}

\begin{figure}[!ht]
    \centering
    \includegraphics[width=0.95\linewidth]{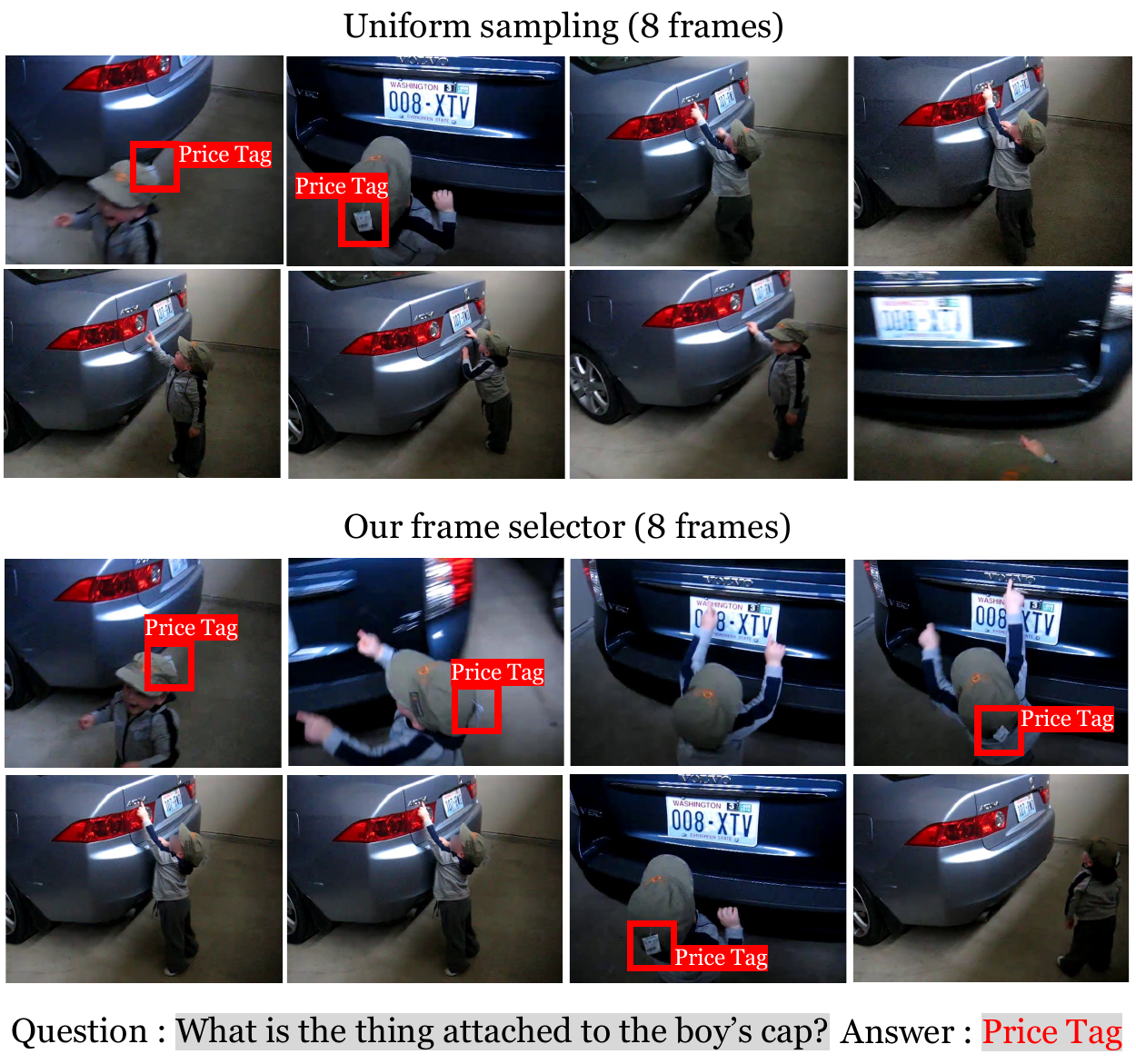}
    \caption{An example of our video frame selection for video QA. Compared to uniform sampling, ours has higher hit rate.}
    \label{fig:head}
    \vspace{-12pt}
\end{figure}

\acf{llm} has revolutionized numerous domains in \acf{ai}~\citep{openai2022chatgpt, claude_technical_report, team2023gemini, touvron2023llama2}. In the past year, \acf{mllm} has significantly improved the performances of \acf{vlm} to an unprecedented level in tasks such as image captioning and \acf{vqa}~\citep{alayrac2022flamingo, li2024llava, ataallah2024minigpt4, Qwen2VL}. To extend \ac{vqa} to the temporal domain, video QA task requires a model to understand consecutive images along the time to answer a question, raising new challenges to the current \ac{mllm}. One of the most critical challenges is the length of the video context, where a model needs to comprehend all frames in a video.

To balance the trade-off between the capability of understanding all video frames and the available context length in a specific \ac{mllm}, conventional practices~\citep{xu2024pllava, zhang2024llavanext-video} rely on uniform sampling of frames. Frames are extracted at pre-defined intervals regardless of their relevance to the specific questions. Although such methods maximize the coverage of a video in time-axis, it introduces insufficient visual information. The uniformly sampled frames may be irrelevant or redundant, meanwhile some important frames are ignored. This fact prevents the model from focusing on key events in the video and sometimes increases the computational costs. Generally in video QA~\citep{yu2019activityqa, liu2024llavanext}, some specific frames are more likely to contain information relevant to the question. The conventional one-size-fits-all approach limits both the performance and practicality of \ac{mllms}, especially when working with long videos or resource-limited environments.

To address these limitations, a straightforward idea is to select frames instead of uniform sampling~\citep{yu2024self, park2024too}. Focusing on the frames that are most helpful for the question, we can use significantly less number of frames without sacrificing the quality of video understanding. Illustrated as an example in Figure \ref{fig:head}, frames that contains the most informative visuals can help downstream model to answer the question. To implement the above idea, we propose a light weight frame selector that employs fine-tuned version of an \ac{llm}. It makes use of \ac{mllm}’s capability of multi-modality understanding to effectively capture the relevance between video frames and the question. We introduce two design choices to make the framework lightweight: (i) small \ac{llm}s are capable to understand complicated user questions; (ii) compress per video frame tokens to balance the long context trade-off. 

Although video QA tasks often require a large number of tokens per frame to capture content details, we hypothesize that determining frame importance does not require excessive tokens. Instead of leveraging all visual tokens of the frames, we apply an aggressive pooling-based token reduction on each video frame, which significantly improves the computational efficiency and increases the number of frames that the lightweight \ac{llm} selector can ingest.

We also propose a training method for our \ac{llm}-based video frame selector. Since well-maintained video frame selection datasets for video QA are scarce, supervised training alone is not feasible. To address this issue, we adopt two pseudo-labeling strategies to estimate frame importance. The first strategy focuses on spatial understanding: we prompt a well-trained \ac{mllm} using a video QA question, which then assigns importance scores to each frame. However, due to the limited context length of a \ac{mllm}, it cannot evaluate all frames together from a temporal viewpoint. Therefore, our second strategy leverages an \ac{llm} to identify the top-k relevant frames using their captions. Instead of visual tokens, the \ac{llm} can analyze more frames simultaneously with caption. We integrate these two approaches to generate pseudo-labels that reflect frame importance relative to a specific question for effective training of the frame selector.

Our frame-selector employs a plug-and-play design that requires no additional fine-tuning of the downstream \ac{mllm}. It reduces noise caused by irrelevant frames, allowing the video-LLM to focus more effectively on the relevant content. Additionally, the selector only needs to be trained once and can subsequently enhance the video QA performance of multiple \ac{mllm}s. We demonstrate significant improvements with several popular \ac{mllms} on video question-answering tasks across various benchmarks, including short-to-medium context (ActivityNet, NExt-QA) and long context (EgoSchema, VideoMME) scenarios. In summary, our contributions are threefold:
\begin{itemize}
    \item We propose a lightweight \ac{mllm}-based adaptive video frame selector to for both efficient and stronger video QA performances of \ac{mllms}.
    \item We propose spatial and temporal pseudo-labeling to generate importance scores for video frame selector training.
    \item Our proposed method is plug-and-play friendly. We demonstrates comprehensive video QA improvements across popular \ac{mllms} with further fine-tuning.
\end{itemize}
\vspace{-3pt}
\section{Related Work}
\vspace{-3pt}
\noindent \textbf{\acf{mllm}} As \acf{llms} continue to demonstrate impressive abilities in language comprehension and reasoning~\citep{achiam2023gpt,team2023gemini,claude_technical_report, touvron2023llama2}, interest is growing within the computer vision community to explore their potential for handling multi-modal inputs. Flamingo~\citep{alayrac2022flamingo} demonstrates the capacity to process image and text inputs for a wide range of multi-modal tasks. BLIP-2~\citep{li2023blip2} introduces a Q-Former to map learned image features into the text embedding space of LLMs, while LLaVA~\citep{liu2023llava} employs simple \acf{mlp} projector to align visual and textual features. Further research has focused on best practices for \ac{mllm}, including areas such as dynamic high-resolution~\citep{liu2024llavanext, chen2024far}, instruction-tuning data~\citep{li2023mvbench, liu2024points}, and different visual encoders~\cite{tong2024cambrian, wei2025vary}. MM1~\citep{mckinzie2024mm1} and Idefics2~\citep{laurenccon2024matters} provide comprehensive ablation studies on the design space of \ac{mllm}.

\begin{figure*}[!ht]
    \centering
    \includegraphics[width=0.87\linewidth]{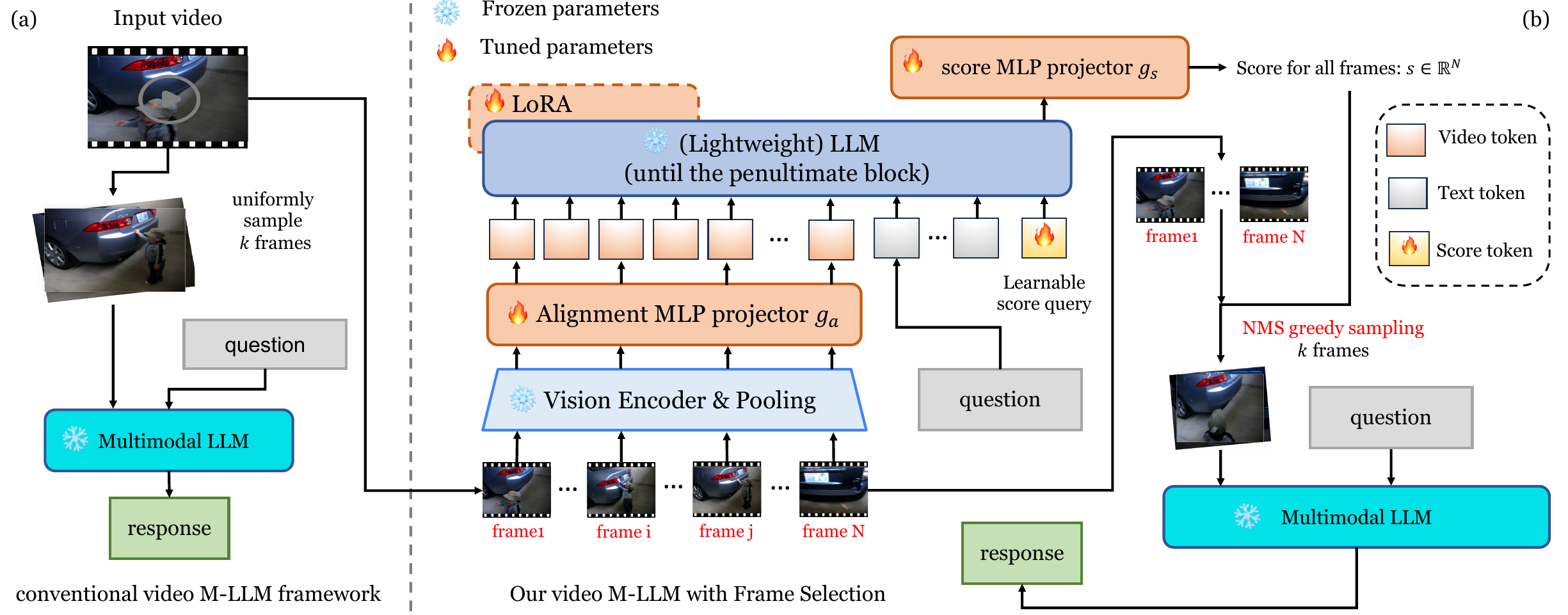}
    \caption{An illustration of the conventional \emph{n}-frame video \ac{mllm} framework and our video \ac{mllm} framework with frame selection.}
    \vspace{-9pt}
    \label{fig:pipe}
\end{figure*}

\noindent \textbf{Video \ac{mllm}s} As image-based \ac{mllm} become more mature, research naturally extends to the video modality. Video-ChatGPT~\citep{Maaz2023VideoChatGPT} and Valley~\citep{luo2023valley} use pooling over video features to generate compact visual tokens for downstream \ac{mllm}. Video-LLaVA~\citep{lin2023video} aligns images and videos before projection, allowing \ac{llm} to learn from a unified visual representation. Video-Teller~\citep{liu2023video} points out the importance of modality alignment in pre-training. PLLaVA~\citep{xu2024pllava} studies how different pooling of visual features affects downstream video question answering performance. Regarding long video inputs, LLaMA-VID~\citep{li2025llama} represents each frame with two tokens to reduce the overload of long videos while preserving critical information. MovieChat~\cite{song2023moviechat} propose an effective memory management mechanism to reduce the computation complexity and memory cost, enabling very long video with over 10K frames understanding. \citet{weng2025longvlm} proposes to extract video representations as sequences of short-term local features, and integrate global semantics into each short-term segment feature.

\noindent \textbf{Video Frame Selection} Before the rise of video \ac{mllm}, language-aware video key-frame selection and localization had attracted great interest~\citep{gao2022mist,liu2022ts2,wang2022contrastive,cui2022video}. \citet{buch2022revisiting} optimized an end2end pipeline that uses ground truth question-answering labels to select a single key frame for downstream tasks. \citet{lu2022lgdn} uses a CLIP-like paradigm to train a transformer for visual and text feature alignment. \citet{qian2022locate} trains a video clip proposal model and the downstream QA model in a iterative manner. \citet{kim2023semi} employs a semi-parametric retriever to obtain the key frames by comparing similarities between frame and language features.

\noindent \textbf{Video \ac{mllm} Frame Selection} Several works propose to select key frames to improve video QA performance. SeViLA~\citep{yu2024self} prompts an \ac{mllm} to obtain a relevance score to each frame, and uses these scores to localize important frames. MVU~\citep{ranasinghe2024understanding} and adopts a similar approach as SeViLA. However, a critical limitation of this kind of methods \citep{han2024self,liang2024end} is the absence of temporal reasoning. Each frame is assessed independently without contextual information from other frames. Furthermore, this method is expensive during inference. Importance score of every frame is estimated via a \ac{mllm}. The longer the video the higher the cost. In contrast, our method outputs importance scores for all frames in a single pass with the compressed visual tokens, which reduces the computational costs and enables temporal reasoning. Koala~\citep{tan2024koala} uses sparsely sampled key frames as conditions for processing subsequent visual tokens. ViLA~\citep{wang2024vila} trains an end-to-end frame selection module to mask input frames. However, both of these methods lack text awareness: during inference, Koala's subsequent visual token processing and ViLA's frame masking do not account for the specific question posed about the video. Recent work such \citep{wang2024videotree, wang2024videoagent} are not end-to-end methods. ~\citet{han2024videoespresso} finds frame selection is helpful for video chain-of-thought reasoning.
\vspace{-3pt}
\section{Method}
\vspace{-3pt}

This section introduces our video frame selector designed for efficient video-LLM QA. Section~\ref{sec:method.1} disccuss our motivation. Section~\ref{sec:method.2} outlines the design details of the frame selector. Section~\ref{sec:method.3} explains the generation of pseudo labels for training the frame selector. Section~\ref{sec:method.4} describes the training process of the frame selector.

\vspace{-3pt}
\subsection{Rethinking Uniform Sampling in Video LLMs}~\label{sec:method.1}

\vspace{-15pt}
\noindent\textbf{A typical framework for video LLM} \; An \emph{n-frame} framework is widely adopted in existing research in video \ac{mllm}~\cite{li2023videochat, xu2024pllava, zhang2024llavanext-video}. The number of frames in the input video, denoted by $T$, is variable. For example, a 3-minute video at 30 \acf{fps} contains $T = 5400$ frames. The \emph{n-frame} framework uniformly samples a fixed number of frames, $[x_1, x_2, \cdots, x_n]$, from the total $T$ frames, where each frame $x_i \in \mathbb{R}^{H \times W \times 3}$, with $H \times W$ representing the frame resolution, and typically $n \ll T$. A pre-trained visual encoder $f_v$ extracts visual features from \emph{n} frames. These features are subsequently projected into the \ac{llm}'s space using an alignment projector $g_a$ and then flattened. Spatial pooling may also be applied to reduce the number of output tokens:

\vspace{-6pt}
\begin{equation}
    h_i = \textit{AvgPooling}(g_a(f_v(x_i))), h_i\in\mathbb{R}^{m\times d}
\end{equation}
where $m$ is the number of tokens to represent a frame and $d$ is the hidden dimension of the \ac{llm}. Let $Q \in \mathbb{R}^{l \times d}$ denote the input embedding of the input text question. The \emph{n-frame} framework generates a response $r$ as following:

\vspace{-6pt}
\begin{equation}
    r = \textit{LLM}(h_1, \cdots, h_n, Q).\label{eq:n-frame}
\end{equation}

\noindent \textbf{Uniform sampling is not optimal}\; In the \emph{n-frame} framework, as shown in Figure~\ref{fig:pipe} (a), the input video is represented by $n \times m$ tokens where $m$ is the number of visual tokens per frame. For example, in LLaVA-NeXT-Video~\citep{zhang2024llavanext-video}, where $n=32$ and $m=12\times12$, this results in 4608 tokens. To reduce the computational cost of \ac{llm} inference, previous work has either chosen to reduce $n$ with sliding Q-Former~\citep{li2023videochat} or $m$~\cite{xu2024pllava, li2025llama} by spatial pooling, in other words, reducing $m$. However, both of them ignore the importance of reducing $n$, the number of intake frames, before encoding and $n$ could be large especially in long videos. Denser uniform sampling makes a larger $n$, thereby reducing the efficiency of the video \ac{mllm}. On the other hand, sampling fewer frames risks omitting crucial information. For example, sampling 32 frames from a 3-minute video means taking one frame every 6 seconds, potentially missing actions that occur within shorter time windows. In fact, most questions about a video can be answered using only a limited number of key frames. Inspired by this factor, we argue that adaptive frame selection according to question is more efficient and effective than uniform frame sampling.

\vspace{-3pt}
\subsection{Design of the Frame Selector}\label{sec:method.2}
\vspace{-3pt}
An ideal frame selector should be able to understand complex questions and analyze temporal relevance in the video input. To achieve this, we fine-tune an \ac{llm} to function as the frame selector. The frame selector utilizes the base \ac{llm}’s strong language comprehension and reasoning abilities to identify key frames in a video.

Similar to existing video \ac{mllm}, our \ac{mllm}-based frame selection method takes as inputs $n$ sampled video frames along with the corresponding text-based question. Instead of generating an answer to the question, the frame selector identifies the most relevant frames for answering the question. Specifically, we make use of well-trained decoder only \ac{llm} as the frame selector to output an $n$-dimensional vector $s$ as follows:

\vspace{-6pt}
\begin{equation}
    s = \textit{FrameSelector}(x_1, \cdots, x_n, Q)\in \mathbb{R}^n
\end{equation}
where the $i^\text{th}$ element in the vector $s$ indicates the importance score of the $i^\text{th}$ input frame. To achieve this, we append a learnable score query $q\in\mathbb{R}^{1\times d}$ to the end of all input tokens and use the concatenation of the visual tokens, text tokens and the score token as the LLM as input:
\begin{equation}
    e_1, \cdots, e_n, e^Q, e^q = \textit{LLM}(x_1, \cdots, x_n, Q, q_{score})
\end{equation}
where $e_i$ and $e^q$ denote the intermediate output of $x_i$  and $q_{score}$ from the penultimate transformer block. Because the nature of causal attention, $e^q$ aggregates information from all visual and text tokens. We then employ an \ac{mlp} to exact frame importance information from the intermediate output of the score query from the penultimate transformer block:

\vspace{-6pt}
\begin{equation}
    s = \textit{MLP}(e^q), s\in\mathbb{R}^n
    \vspace{-12pt}
\end{equation} \label{eq:selector}

Figure~\ref{fig:pipe} (b) presents an overview of the proposed architecture for the \ac{mllm}-based frame selector. Instead of generating tokens that represent the frame selection, we append a learnable query vector at the end of the input sequence and supervisedly learn this query token. The hidden vector of this query token, $e^q$, then serves as the input to generate the $n$-dimensional importance vector $s$.

\begin{algorithm}[t]
    \caption{Greedy NMS sampling}\label{alg:1}
    \begin{algorithmic}[1]
        \State \textbf{Input:} Importance score $s\in\mathbb{R}^n$, Number of frames selected $k$
        \State Initialize the neighbor gap $\delta\gets\text{integer}(\nicefrac{n}{4k})$.
        \State Initialize the selected index list $I_s=[]$.
        \For{step in $1\cdots, k$} 
        \State (Greedy) Selected frame index $i\gets \arg\max s.$ Append index $i$ to $I_s$.
        \State (NMS) Update the importance score:\\ $s[j] = -1 \;\text{if}\; |i-j|\leq \delta$
        \EndFor
        \State $I_s\gets \text{sort}(I_s)$
        \State \textbf{Return:} $I_s$
    \end{algorithmic}\label{alg2}
\end{algorithm}

\begin{figure*}[t]
    \centering
    \includegraphics[width=0.83\linewidth]{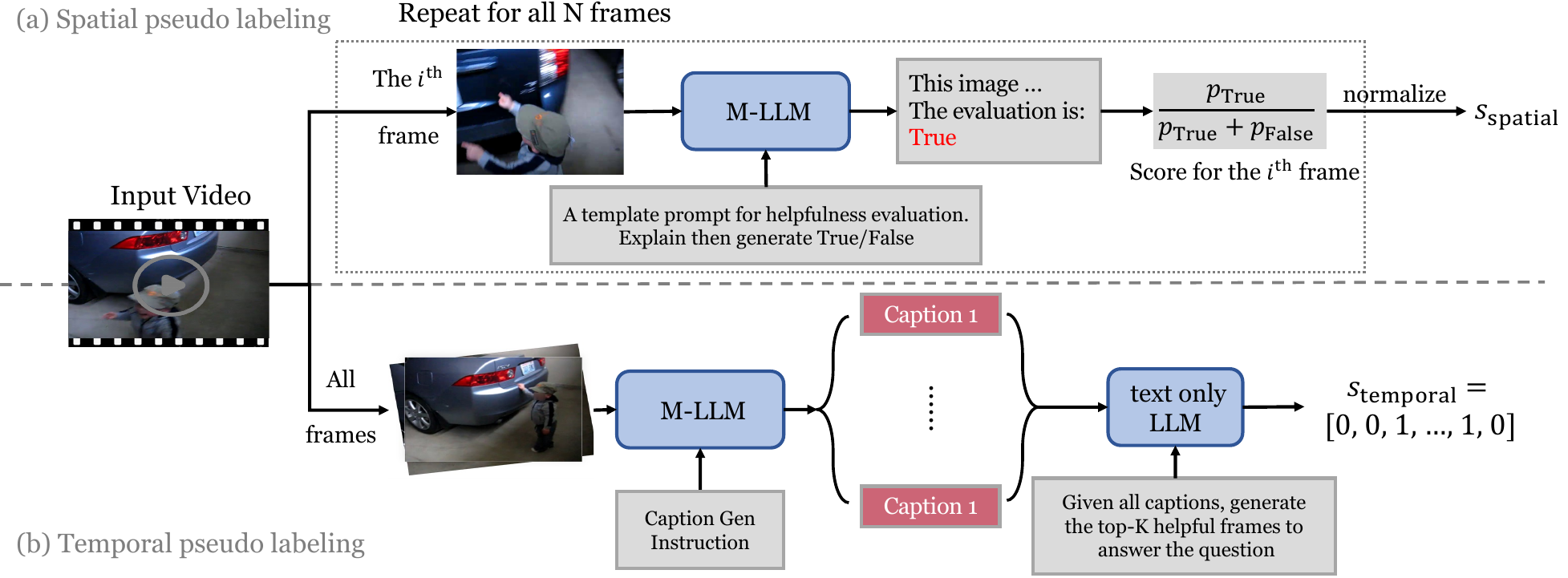}
    \caption{An illustration of the spatial and temporal pseudo labeling for the importance scores}
    \vspace{-12pt}
    \label{fig:pseduo_label}
\end{figure*}

\noindent \textbf{Select frames from the importance score} After obtaining per-frame importance scores, we need to sample $k$ frames for downstream video question-answering via a video \ac{mllm}. Naively selecting the top $k$ frames with the highest importance scores is suboptimal because neighboring frames within short time intervals often have similar scores. For example, if frame $i$ has the highest importance score, the $i+1$ or $i-1$ frames usually have a closely high scores. The adjacent frame adds little additional information if frame $i$ is already selected. To address this issue, we use a greedy algorithm combined with non-maximum suppression (NMS) to select the most informative frames. The ``Greedy'' approach involves sequentially selecting the top k frames, without replacement, by choosing the frame with the highest importance score from the remaining set of frames. With ``NMS'', once a frame is selected, its neighboring frames are excluded as they contain similar information. The ``NMS-Greedy'' procedure is detailed in Algorithm~\ref{alg:1}. In practice, when selecting $k$ frames from $n$ frames, frame $j$ is considered a neighboring frame of frame $i$ if $|i-j| \leq (\nicefrac{n}{4k})$.

\noindent \textbf{Efficiency of the frame selector}\; As discussed in Section~\ref{sec:method.1}, dense uniform sampling is inefficient for video \ac{mllm}. However, we still employ dense uniform sampling for the frame selector.

Although dense uniform sampling is inefficient for video \ac{mllm}, we keep using it at the beginning to maximally preserve the video information. To overcome the drawback of dense uniform sampling, we apply spatial pooling to reduces the token count per video frame before feeding the frames into the selector. In particular, the spatial pooling reduce the number of visual token to a smaller value, e.g., $m = 3 \times 3$ (9 tokens), which is substantially fewer than the tokens per frame used in existing video LLMs, e.g., $m = 12 \times 12$ (144 tokens). This design follows such an assumption: for video question answering, the model requires a substantial number of tokens per frame to capture visual details, but far fewer tokens are needed to determine whether a frame is important. A rough outline of the frame is sufficient. Our empirical results Table~\ref{tab:ab5.1} also justify this assumption.

\vspace{-3pt}
\subsection{Pseudo Labels for the Frame Selector}\label{sec:method.3}
\vspace{-3pt}

To train the frame selector, we need supervision signals for the output importance scores $s\in\mathbb{R}^n$. Unfortunately, there is no existing dataset to label the frame level importance score for video QA, we propose two methods to generate pseudo labels for training the frame selector.

\noindent \textbf{Spatial Pseudo Labels} 
Previous works use \ac{mllms} to evaluate whether a video frame is relevant to a question~\citep{yu2024self, ranasinghe2024understanding}. A common practice is to prompt an \ac{mllm} and ask if the video frame provides useful information to answer the question. The relevant score of the frame is the probability that the \ac{mllm} generates the ``Yes'' token. In our experiments, we observe that this method does not always provide a reasonable estimation. Even if the model agrees the frame is relevant, the \ac{mllm} may not generate the ``Yes'' token but other expressions depending on its text generation style.
To address this issue, we apply chain-of-thoughts (CoT), asking the \ac{mllm} to explain first then generate a Boolean evaluation (check appendix for the detailed prompt). This prompts allows the \ac{mllm} to improve the evaluation quality by extra inference time reasoning. An ideal \ac{mllm} should adhere to the instruction by generating either ``True'' or ``False''. In a few cases the model may fail to follow the instruction, and we manually append the text ``Evaluation: True'' to the end of the generated response. Thus we can always compute the probabilities of generating ``True'' and ``False'', denoted as $p_{\text{True}}$ and $p_{\text{False}}$ respectively. The importance score of the input frame is determined by

\vspace{-9pt}
\begin{equation}
    s = p_{\text{True}}/(p_{\text{True}} + p_{\text{False}})\label{eq:spl}
\end{equation}

In our experimental setup, we uniformly sample $n=128$ frames from the video and obtain the spatial pseudo labels for each frame independently. Let $s_i$ denote the score for the $i^\text{th}$ frame, we normalize the score vector as $ \nicefrac{s_i}{\max_j s_j}$. Figure~\ref{fig:pseduo_label} (a) shows the pipeline.

\vspace{3pt}
\noindent \textbf{Temporal Pseudo Labels} A significant limitation of single-frame evaluation is the lack of temporal reasoning. For example, considering the question: ``\textbf{What did the man do after picking up his hat?}'', the video content following the action of picking up the hat is crucial for answering this question. However, when generating the spatial pseudo labels, it only considers one frame without taking care of the temporal context. Therefore, the spatial labels don't know what occurs after the action.

To address this issue, we propose the temporal pseudo labeling. Since most of the publicly available \ac{mllm}s are not able to consume a large number of image tokens, we alter to take advantages of the frame captions and use a \ac{llm} to reason over all captions. Specifically, we first obtain detail captions of all $n$ frames by prompting a \ac{mllm}. Second, we feed the captions of all frames together as well as the question to a strong text-only \ac{llm}. Then the \ac{llm} can temporally reason the helpfulness of all frames.

We find it challenging to generate floating-point scores for an extensive list of frames for an \ac{llm}. Consequently, we ask the model to produce a list of the index of most helpful frames (see appendix Section \ref{supp:sec:prompt} for details). Frames included in this list are assigned a score of 1, while those excluded receive a score of 0. Figure~\ref{fig:pseduo_label} (b) shows the pipeline.

While temporal pseudo labels can capture temporal relations in videos, it may suffer from information loss and model hallucination due to its two-stage evaluation process. Therefore, we combine two methods into the final pseudo-labels by averaging the scores obtained from spatial and temporal pseudo labels.

\vspace{-3pt}
\subsection{Training of the Frame Selector}\label{sec:method.4}
\vspace{-3pt}
We consider a two-stage training procedure. In stage 1, we freeze the pre-trained vision and LLM backbones and train the parameters of the alignment projector $g_a$, the learned score query $X_\text{score}$ and the score projector $g_s$. Figure~\ref{fig:pipe} (b) shows the trainable modules in the red boxes and the frozen modules in the blue boxes. Stage 1 training is optimized over the two tasks alternatively. We use below two tasks:

\begin{itemize}
    \item \textbf{Visual instruction following}  Recall $r$ in Equation~\ref{eq:n-frame} is the generated response from the LLM, the objective is the cross entropy loss between $r$ and the  ground truth response. This tasks trains the projector $g_a$ to aligns the visual features with the pre-trained LLM embedding space.
    \item \textbf{Importance score prediction} Recall $s$ in Equation~\ref{eq:selector} is the importance score for $n$ frames, the objective is the binary cross entropy loss between $s$ and the pseudo labels generated from Section~\ref{sec:method.3}. This task provides a good initialization for the score query and the score projector.
\end{itemize}

In stage 2, we only train the model with the importance score prediction task. Besides the alignment projector $g_a$, the learned score query $q$ and the score projector $g_s$, we also include the \ac{lora} weights of the \ac{llm} as the trainable parameters to adapt the \ac{llm} to the frame selection task.

\vspace{-6pt}
\section{Experiments}
\vspace{-3pt}

We begin by outlining our experimental setup in Section~\ref{sec:exp.1}. Then we demonstrate that our frame selector improves the performance of well-trained \ac{mllms} without changing their parameters by selecting better frames in Section~\ref{sec:exp.2}. We also conduct ablation studies to demonstrate the effectiveness and efficiency of our frame selection framework and results are in Section~\ref{sec:exp.3}. At the end, we showcase some qualitative examples of frames selected from the video according to the question in Section~\ref{sec:exp.4}.

\vspace{-3pt}
\subsection{Experiment Setup}~\label{sec:exp.1}
\vspace{-15pt}

\noindent\textbf{Training data} We compile the training dataset from three sources: 1) 800K data from VideoChat2~\citep{li2024mvbench}, 2) 125K data from the TimeIT dataset~\citep{ren2024timechat}, and 3) 178K data from the LLaVA-Video-178K dataset~\citep{zhang2024videoinstructiontuningsynthetic}. For the visual instruction tuning task, we use the entire training dataset. For the importance score prediction task, we use 400K video QA data where the video length exceeds 5 seconds.
\noindent\textbf{Pseudo label generation} we utilize Qwen2-VL-7B~\citep{Qwen2VL} to generate importance scores for each frame in the  spatial pseudo labeling and concise captions for all frames in the temporal pseudo labeling. We use GPT-4o mini to propose the most helpful frames given the concise captions in the multi-frame evaluation.
\noindent\textbf{Evaluation benchmarks} Since our framework selects frames for video question answering, we evaluated its performance on benchmarks consisting of relatively longer videos, including open-ended video QA on ActivityNet-QA~\citep{yu2019activitynet}, and multi-choice QA on NExT-QA~\citep{xiao2021next}, VideoMME~\citep{fu2024video}, EgoSchema~\citep{mangalam2023egoschema}, LongVideoBench~\citep{wu2024longvideobench}.

\noindent \textbf{Implementation details} We use the pre-trained SigLIP ViT-Large~\citep{zhai2023sigmoid} as the visual encoder and Qwen2.5 1.5B~\citep{qwen2.5} as the backbone of the \ac{llm}. We uniformly sample 128 frames from the video as the input for the visual encoder. The output from the visual encoder have a size of $128 \times 16 \times 16$. After the alignment projector and the spatial pooling layer, the size of the visual tokens is $128 \times 3 \times 3$. In Stage 1, we use a batch size of 128, a learning rate of $10^{-3}$, and a warm-up ratio of 0.03 to train the model for two epochs. In Stage 2, a batch size of 128, a learning rate of $10^{-5}$, a warm-up ratio of the first 3\% iterations, and a cosine learning rate scheduler are used to train the model for five epochs. 

\begin{table}[htbp]
    \centering
    \resizebox{\linewidth}{!}{
        \begin{tabular}{lrr}
        \toprule
        Model          & Model Size &ActivityNet-QA \\ \midrule
        Video-ChatGPT \citep{Maaz2023VideoChatGPT}& 7B   & 35.2 / 2.8  \\
        Chat-UniVi~\citep{jin2024chat}& 7B   &     46.1 / 3.3  \\
        LLaMA-VID~\citep{li2025llama}& 7B   &      47.4 / 3.3  \\
        LLaMA-VID~\citep{li2025llama} & 13B  &      47.5 / 3.3 \\
        Video-LLaVA \citep{lin2023video} & 7B   &    45.3 / 3.3  \\
        MiniGPT4-Video~\citep{ataallah2024minigpt4}& 7B   &  46.3 / 3.4   \\
        SlowFast-LLaVA~\citep{xu2024slowfast}& 7B   &    55.5 / 3.4  \\
        SlowFast-LLaVA~\citep{xu2024slowfast}& 34B  &     59.2 / 3.5 \\
        Tarsier~\citep{wang2024tarsier}& 7B   & 59.5 / 3.6\\
        Tarsier~\citep{wang2024tarsier}& 34B  & 61.6 / 3.7  \\
        PLLaVA~\cite{xu2024pllava}         & 7B   & 56.3 / 3.5 \\
        PLLaVA~\cite{xu2024pllava}         & 34B  & 60.9 / 3.7  \\
        LLaVA-NeXT-Video~\citep{zhang2024llavanextvideo} & 7B        &  53.5 / 3.2   \\
        LLaVA-NeXT-Video~\citep{zhang2024llavanextvideo} & 34B        &  58.8 / 3.4   \\
        \midrule
        PLLaVA + Selector & 7B + 1.5B &  57.6 (1.3$\uparrow$) / 3.5  \\
        PLLaVA + Selector & 34B + 1.5B & 62.3 (1.4$\uparrow$) / 3.6  \\
        LLaVA-NeXT-Video + Selector & 7B + 1.5B &  55.1 (1.6$\uparrow$) / 3.4\\
        LLaVA-NeXT-Video + Selector & 34B + 1.5B & 60.2 (1.4$\uparrow$) / 3.5  \\
        \bottomrule
        \end{tabular}
    }
    \vspace{-6pt}
    \caption{Comparison of open-ended question answering evaluation on ActivityNet QA. Results with the ``\textbf{+ Selector}'' are ours.}
    \label{tab:anetqa}
    \vspace{-9pt}
\end{table}

\vspace{-3pt}
\subsection{Comparison with SOTA Video-LLMs}~\label{sec:exp.2}
We choose two strong video \ac{mllm}s, PLLaVA~\citep{xu2024pllava} and LLaVA-NeXT-video~\citep{zhang2024llavanextvideo} and two (multi-)image based \ac{mllm} Idefics~\citep{laurenccon2024matters} and Qwen2-VL~\citep{Qwen2VL}, to be the baselines to illustrate how our frame selector enhances the video question-answering (QA) performance of these models. For each model, we compare the performance of using uniformly sampled frames as inputs versus using frames selected by our frame selector. The number of frames seen by the video \ac{mllm} is the same during the comparison.

\begin{table}[htbp]
\centering
\resizebox{\linewidth}{!}{
\begin{tabular}{lrr}
\toprule
Model          & Model Size & NExT-QA\\ \midrule
SlowFast-LLaVA~\citep{xu2024slowfast}& 7B   &    64.2 \\
SlowFast-LLaVA~\citep{xu2024slowfast}& 34B  &     72.0 \\
Tarsier~\citep{wang2024tarsier}& 7B   & 71.6 \\
Tarsier~\citep{wang2024tarsier}& 34B  & 79.2 \\
LLaVA-NeXT-Video~\citep{zhang2024llavanextvideo} & 7B        &  62.4 \\
LLaVA-NeXT-Video~\citep{zhang2024llavanextvideo} & 34B   & 68.1\\ 
Idefics2~\citep{laurenccon2024matters} & 8B & 68.0 \\
Qwen2-VL~\cite{Qwen2VL} & 7B  & 77.6   \\\midrule
LLaVA-NeXT-Video + Selector & 7B + 1.5B &  63.4 (1.0$\uparrow$) \\
LLaVA-NeXT-Video + Selector & 34B + 1.5B & 69.3 (1.2$\uparrow$) \\
Idefics2 + Selector  & 8B + 1.5B & 69.1 (1.1$\uparrow$) \\
Qwen2-VL + Selector & 7B + 1.5B  &78.4 (0.8$\uparrow$)    \\
\bottomrule
\end{tabular}
}
\vspace{-6pt}
\caption{Comparison of multi-choice question answering evaluation on NExT-QA. Results with the ``\textbf{+ Selector}'' are ours.}
\vspace{-10pt}
\label{tab:nextqa}
\end{table}

\begin{table}[htbp]
\centering
\resizebox{\linewidth}{!}{
\begin{tabular}{lrr}
\toprule
Model          & Model Size &  EgoSchema\\ \midrule
SlowFast-LLaVA~\citep{xu2024slowfast}& 7B   &    47.2 \\
SlowFast-LLaVA~\citep{xu2024slowfast}& 34B  &      55.8 \\
Tarsier~\citep{wang2024tarsier}& 7B   & 56 \\
Tarsier~\citep{wang2024tarsier}& 34B  & 68.6 \\
LLaVA-NeXT-Video~\citep{zhang2024llavanextvideo} & 7B        &  45.8 \\
LLaVA-NeXT-Video~\citep{zhang2024llavanextvideo} & 34B   &  48.6 \\ 
Idefics2~\citep{laurenccon2024matters} & 8B& 56.6 \\
Qwen2-VL~\cite{Qwen2VL} & 7B  &  64.6  \\\midrule
LLaVA-NeXT-Video + Selector & 7B + 1.5B  & 47.2 (1.3$\uparrow$) \\
LLaVA-NeXT-Video + Selector & 34B + 1.5B& 50.6 (2.0$\uparrow$)  \\
Idefics2 + Selector  & 8B + 1.5B & 57.9 (1.3$\uparrow$) \\
Qwen2-VL + Selector & 7B + 1.5B & 65.9 (1.1$\uparrow$)   \\
\bottomrule
\end{tabular}
}
\vspace{-6pt}
\caption{Comparison of multi-choice question answering evaluation on EgoSchema. Results with the ``\textbf{+ Selector}'' are ours.}
\label{tab:egoscheaeq}
\end{table}

\begin{table}[htbp]
    \centering
    \small\addtolength{\tabcolsep}{-3pt}
    \begin{tabular}{@{}lllll@{}}
    \toprule
    Model            & short & medium & long & average  \\ \midrule
    Qwen2-VL (baseline) & 69.1  & 53.0   & 51.6 & 58.1 \\
    Qwen2-VL + our selector & 69.6 & 54.1 & 51.9 & 58.7 (0.6$\uparrow$) \\ \bottomrule
    \end{tabular}
    \vspace{-6pt}
    \caption{Performances on VideoMME. \textbf{``+ Selector''} are ours.}
    \vspace{-12pt}
    \label{tab:rebuttal_longvideo_result}
\end{table}

Table~\ref{tab:anetqa}, Table~\ref{tab:nextqa}, Table~\ref{tab:egoscheaeq} and Table~\ref{tab:rebuttal_longvideo_result} present the comparison on ActivityNet-QA~\citep{yu2019activitynet}, NExT-QA~\citep{xiao2021next} and EgoSchema~\citep{mangalam2023egoschema}, VideoMME~\citep{fu2024video} respectively. Following existing practice~\citep{Maaz2023VideoChatGPT}, performance on ActivityNet-QA is measured as ``accuracy/correctness'' metrics, and both metrics are evaluated using GPT-3.5, with higher values indicating better performance. Performances on NExT-QA and EgoSchema are the accuracy of of multi-choice questions where each question has 5 options. We prefill the word ``Option'' as the initial token to generate, and then use the next generated token as the prediction result.

\subsection{Ablation Studies}~\label{sec:exp.3}
We conduct ablation studies on several components of the frame selection framework. First, we analyze the performance of the following frame selection methods in the video QA tasks:

\begin{itemize}
    \item \textbf{Uniform sampling}: the default uniform sampling.
    \item \textbf{Pseudo labels from CLIP similarity}: define the importance score of a frame as the image-text similarity between the frame and the text question, computed using a CLIP model. Then sample frames using Algorithm~\ref{alg:1}.
    \item \textbf{Pseudo labels from SeViLA}: define the importance score of a frame as in SeViLA~\citep{yu2024self} to sample frames.
    \item \textbf{Spatial pseudo labels}: define the importance score using the spatial pseudo labels only.
    \item \textbf{Spatial \& temporal pseudo}: define the importance score as the average of the spatial and temporal pseudo labels.
    \item \textbf{Trained selector}: define the importance score of a frame using the output of our trained frame selector.
\end{itemize}

\noindent\textbf{Not using video-LLM due to worse pseudo-label quality.} We refer to \textit{Table 3} in \cite{fang2024mmbench} for justification. In the temporal reasoning tasks, although video-LLMs are trained with frames sampled from the same video, they fall behind image-trained M-LLMs. It implies image-trained M-LLMs could lead to better temporal pseudo-labels for our method. We tried to use LLaVA-NeXT-Video to generate pseudo-labels. It is more time-consuming but worse pseudo-labels quality. On NExT-QA, using frames labeled by Qwen2-VL and LLaVA-NeXT-Video are 63.9\% and 62.8\% respectively.

\begin{table}[htbp]
    \centering
    \resizebox{\linewidth}{!}{
    \begin{tabular}{lrr}
    \hline
    Selection Method & ANet-QA  &  NExT-QA \\
    \hline
    Uniform sampling                & 53.5 & 62.4 \\
    Scores from CLIP similarity     & 53.7 & 62.2 \\
    Pseudo labels from SeViLA       & 54.0 & 63.2 \\
    Spatial pseudo labels & 54.2 & 63.6 \\
    Spatial \& temporal pseudo  & 55.5 & 63.9 \\
    Scores from trained selector    & 55.1 & 63.4 \\
    \hline
    \end{tabular}
}
\vspace{-6pt}
\caption{Performance of LLaVA-NeXT-Video 7B on ActivityNet (ANet) and NExT QA with different frame selection methods}
\vspace{-6pt}
\label{tab:ab2}
\end{table}

Table~\ref{tab:ab2} presents the video QA performance of LLaVA-NeXT-Video 7B on two benchmarks, respectively. \textbf{Uniform} and \textbf{CLIP} sampling serve as simple baselines for frame selection, while other methods utilize \ac{mllm} reasoning. Both \textbf{SeViLA} and our \textbf{Spatial} are single-frame-based selection methods, with the latter achieving superior performance due to enhanced reasoning during multimodal LLM inference. \textbf{Spatial \& Temporal} further improves upon \textbf{Spatial}, demonstrating the importance of temporal reasoning in frame selection. However, the computational cost to generate such pseudo labels are extremely high as it needs to prompt an \ac{mllm} densely. The light-weight selector's performance matches the performance of frame selection using pseudo labels, validating the effectiveness of the selector architecture.

\begin{table}[]
\centering
\resizebox{\linewidth}{!}{
\begin{tabular}{cccccc}
\toprule
\# frames          & Acc@C & Acc@T & Acc@D & Acc  & Speed (s) \\ \midrule
$4$                & 67.2  & 61.2  &  73.9 & 66.4 & 0.56  \\
$8$                & 68.7  & 62.5  &  76.9 & 68.1 & 0.92  \\
$16$               & 69.1  & 63.6  &  76.8 & 68.7 & 1.71  \\
$32$               & 69.5  & 64.3  &  78.4 & 69.3 & 3.40  \\ \midrule
$128\rightarrow4$  & 68.5  & 64.5  &  75.7 & 68.5 & 0.76  \\
$128\rightarrow8$  & 69.3  & 64.9  &  77.5 & 69.3 & 1.12  \\
$128\rightarrow16$ & 69.4  & 64.8  &  78.5 & 69.5 & 1.91  \\
$128\rightarrow32$ & 69.2  & 65.6  &  78.7 & 69.6 & 3.50  \\ \bottomrule
\end{tabular}}
\vspace{-6pt}
\caption{Performance and inference speed of LLaVA-NeXT-Video 34B on NExT-QA with different number of input frames. First 4 rows: uniform sampling, last 4 rows: sampling using the selector.}
\label{tab:ab3}
\vspace{-9pt}
\end{table}

We further show that the video QA system can use fewer frames to reason videos with the help of our frame selector. Table~\ref{tab:ab3} shows the performance of LLaVA-NeXT-Video on NExT-QA taking different number of frames as input and the inference speed of LLaVA-NeXT-Video with different number of input frames. The inference speed was measured with float16 precision using a batch size of 1 on a single A100 GPU, employing the Hugging Face implementation. When taking the same number of frames as input, our framework incurs additional inference costs to select frames using the \ac{mllm} selector. However, this increase in inference time is not significant thanks to the efficient design of the frame selector. Moreover, the frames selected by the \ac{mllm} selector are more useful for answering the question. Therefore, \textbf{the model using a selector with ${n}$-frame input can achieve similar video QA performance as the model without a selector using ${2n}$-frame input}. For example, the configuration $128\rightarrow4$ outperforms 8-frame uniform sampling with a faster inference speed.

\begin{table}[htbp]
    \centering
    \resizebox{\linewidth}{!}{
    \begin{tabular}{cccc}
    \toprule
    \# tokens / frame  & ActivityNet-QA  &  NExT-QA & EgoSchema\\
    \midrule
    no selector & 53.5 & 62.4 & 45.8 \\
    1     & 53.2 & 62.7 & 46.6\\
    9     & 55.1 & 63.4 & 47.2\\
    25    & 55.3 & 63.6 & 47.3\\
    \bottomrule
    \end{tabular}
    }
    \vspace{-3pt}
    \caption{Performance of \textbf{LLaVA-NeXT-Video 7B} with different number of tokens per frame used in the selector.}
    \label{tab:ab5.1}
    \vspace{-8pt}
\end{table}

\begin{table}[htbp]
    \centering
    \resizebox{\linewidth}{!}{
    \begin{tabular}{cccc}
    \toprule
    Backbone size & ActivityNet-QA  &  NExT-QA & EgoSchema\\
    \hline
    no selector & 53.5 & 62.4 & 45.8 \\
    0.5 B  & 53.8 & 62.8  & 46.4\\
    1.5 B  & 55.1 & 63.4  & 47.2\\
    7 B    & 55.5 & 64.0  & 47.9\\
    \bottomrule
    \end{tabular}
    }
    \vspace{-3pt}
    \caption{Performance of LLaVA-NeXT-Video 7B with different size of the selector's base LLM.}
    \label{tab:ab5.2}
    \vspace{-6pt}
\end{table}

As discussed in Section~\ref{sec:method.2}, one advantage of our framework is that the frame selector features a lightweight design. We examines two key hyperparameters that influence the computational efficiency: the number 
of tokens to represent a frame and the size of the base LLM. By default, we use Qwen2.5 1.5b as the LLM backbone and use 9 tokens for video frame representation. Table~\ref{tab:ab5.1} and Tabel~\ref{tab:ab5.2} present the ablation studies examining these two factors. ``no selector'' indicates sample the video frames uniformly.

We further demonstrate the effectiveness of our frame selector for long video question answering (QA). LongVideoBench~\citep{weng2025longvlm} is a recently introduced benchmark for long-context video-language understanding, with an average video duration of 473 seconds. In Table~\ref{abl:long}, we report the performance of LLaVA-Next-Video 34B and Qwen2-VL 7B with different numbers of input frames, using uniform sampling and sampling with our frame selector. For both \ac{mllms}, the performance with $n$ input frames sampled using the selector surpasses that of $2n$ input frames with uniform sampling, demonstrating the effectiveness of the frame selector.

\begin{table}[]
    \centering
    \resizebox{0.9\linewidth}{!}{
    \begin{tabular}{@{}llll@{}}
    \toprule
    model                                 & \# frames & Uniform & Selector \\ \midrule
    \multirow{4}{*}{LLaVA-Next Video 34B} & 4         & 45.3    & 49.5     \\
                                          & 8         & 46.9    & 49.9     \\
                                          & 16        & 48.1    & 49.8     \\
                                          & 32        & 49.7    & 50.0     \\ \midrule
    \multirow{4}{*}{Qwen2-VL 7B}          & 4         & 48.0    & 55.0     \\
                                          & 8         & 50.9    & 56.0     \\
                                          & 16        & 53.6    & 56.5     \\
                                          & 32        & 53.3    & 57.0     \\ \bottomrule
    \end{tabular}}
    \caption{Performance of LLaVA-NeXT-Video 34B Qwen2-VL 7B on LongVideoBench  with different input frames.}
    \label{abl:long}
    \vspace{-12pt}
\end{table}

\subsection{Visualization of selected frames}\label{sec:exp.4}
\begin{figure}[htbp]
    \centering
    \includegraphics[width=\linewidth]{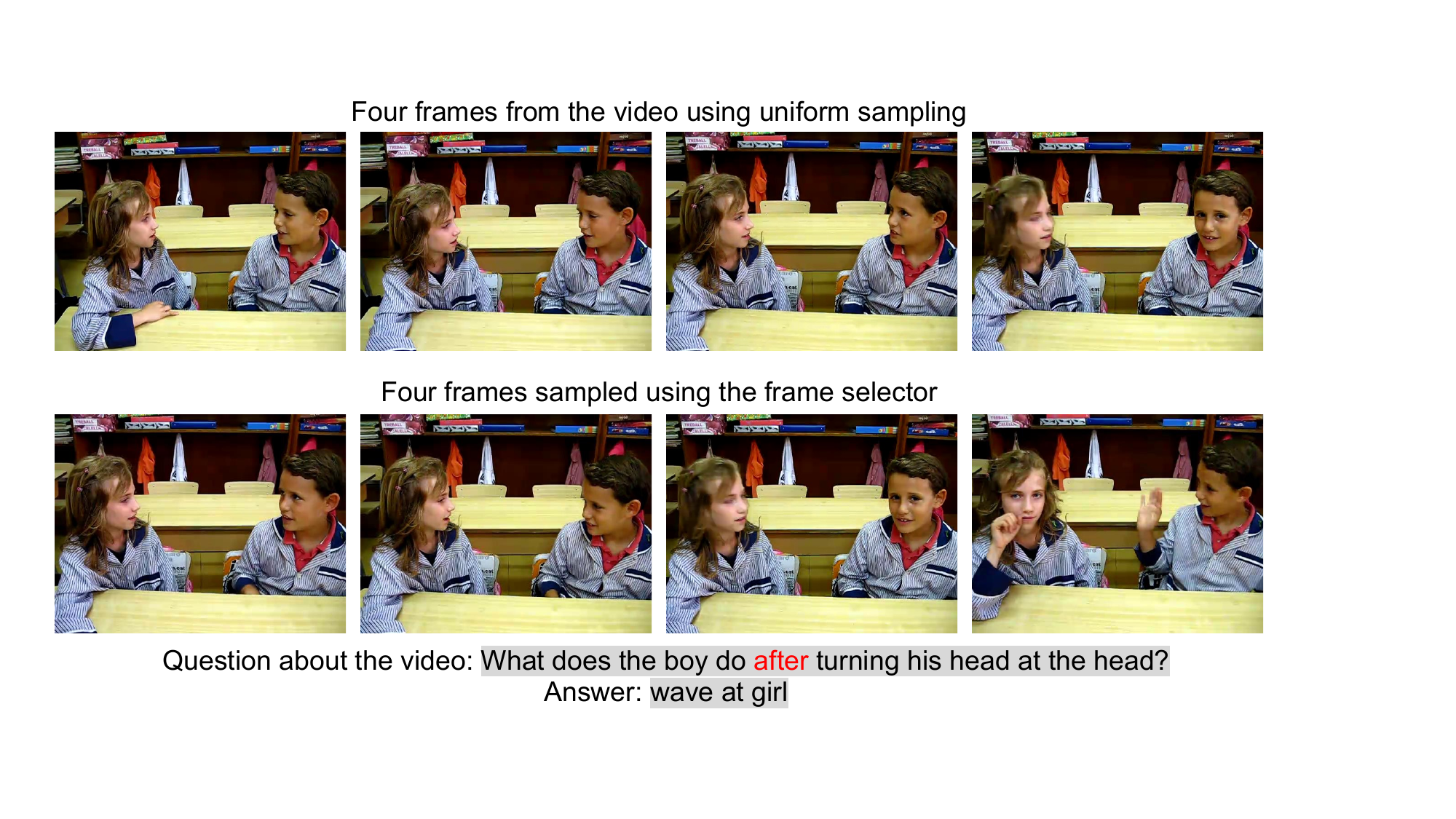}
    \includegraphics[width=\linewidth]{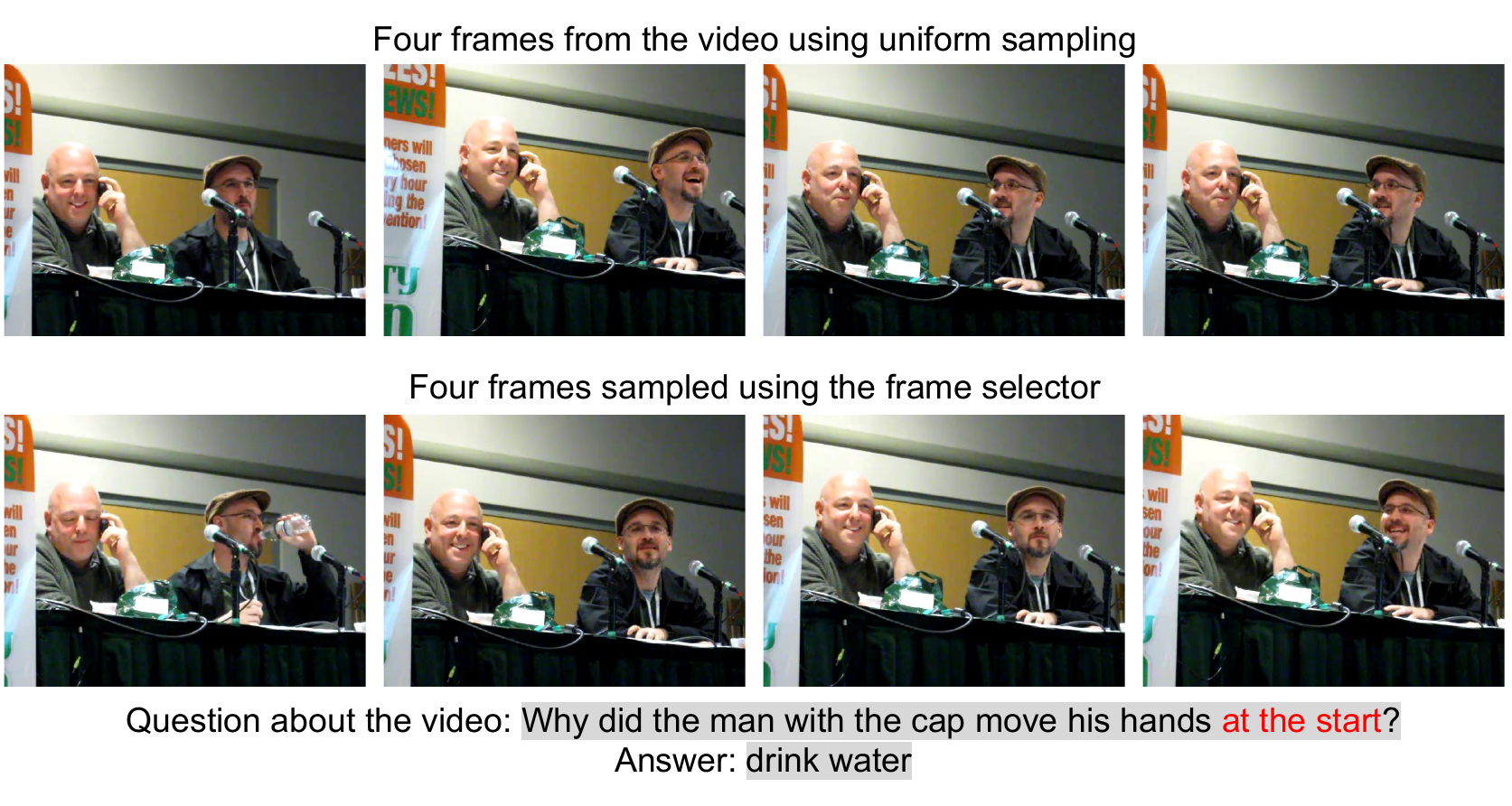}
    \vspace{-9pt}
    \caption{Visualization of the frame selection results.}
    \vspace{-12pt}
    \label{fig:visualazation}
\end{figure}

\noindent\textbf{Qualitative Results:} Figure~\ref{fig:visualazation} presents two examples of selected frames from the video conditioned the question. Each question involves two events and thus requires temporal reasoning. Estimating the importance of a single frame is challenging without reference to prior or subsequent frames. The frame selector effectively identifies frames containing the answers to the questions. More results in Appendix Section \ref{supp:sec:viz} .

\noindent\textbf{Evaluate on Video Grounding Benchmarks:} We compare the moment retrieval performance with SeViLa on QVHighlights. Ours achieves 43.9\% R1@5 and 32.3\% R1@7 while SeViLa achieves 54.5\% R1@5 and 36.5\% R1@7. Although our frame selector has lower performance than SeViLa, it is not specifically trained on QVHighlights. In other words, the comparable performance proves the overall correctness of the frame selection.

\section{Conclusion}
In this paper, we propose a lightweight \ac{mllm}-based frame selector to improve both performances and efficiency in video QA. This selector is question-aware and takes dense video frames and the question as input, selecting the most relevant frames. We can then use any multi-image or video M-LLMs with the selected frames to complete the question-answering. To train the frame selector, we introduce spatial and temporal pseudo-labeling due to the limited public annotations for video frame-level importance. Our experiments on two medium-length video QA benchmarks (ActivityNet QA and NExt-QA) and two long-video QA benchmarks (EgoSchema and LongVideoBench, VideoMME) demonstrate the effectiveness of our proposed method.

\newpage
{
    \small
    \bibliographystyle{ieeenat_fullname}
    \bibliography{references}
}

\appendix
\clearpage
\maketitlesupplementary
\renewcommand{\thesection}{\Alph{section}}

\section{Prompt for Pseudo Label Generation}
\label{supp:sec:prompt}
Table~\ref{tab:spatial} provides the prompt template for generating pseudo spatial labels.
We uniformly sample $n=128$ frames and use the prompt template to obtain a score for each frame independently. We use the logit to generate the word ``True'' or ``False'' after ``Evaluation:'' to compute the score using Equation~\ref{eq:spl}. In a few cases, the \ac{mllm} response may not follow the instruction and does not contain the text ``Evaluation: True'' or ``Evaluation: False''. We manually add `Evaluation: True'' to the end of the response and use the logit to generate the word ``True'' to compute the score.
\begin{table}[ht]
\centering
\begin{tabular}{l}
\toprule
The image is a video frame from a video. A question \\
about the video is:\\
\{question\}\\
Evaluate whether the video frame provides useful \\
information to answer this question about the video.\\
First explain your reasoning. Then generate a\\ Boolean evaluation of the frame's usefulness. For \\
example:\\
Evaluation: True
\\\bottomrule
\end{tabular}
\caption{Prompt template for spatial pseudo labels}
\label{tab:spatial}
\end{table}

\noindent Table~\ref{tab:temporal} provides the prompt template for generating pseudo temporal labels. We first use the \ac{mllm} to generate a concise caption for $n=128$ uniformly sampled frames. Then we use the prompt in Table~\ref{tab:temporal} to generate a list of frame indexes containing the most helpful frames.

\begin{table}[ht]
\centering
\begin{tabular}{l}
\toprule
I need to answer a question based on a long video. To \\
do this, I have uniformly sampled 128 frames from \\
the video, each with a corresponding caption. The question \\
I need to answer is:\\
\{question\}\\
Below is the list of frames and their captions:\\

Frame 1 : \{caption1\} \\
Frame 2 : \{caption2\} \\
$\cdots$\\
Frame 128 : \{caption128\}\\

Please provide a list of 8 frames that would be most \\
helpful for answering this question.\\
Rule: ONLY provide a Python List without extra text.
\\\bottomrule
\end{tabular}
\caption{Prompt template for temporal pseudo labels}
\label{tab:temporal}
\end{table}

\begin{table}[ht]
    \centering
    \resizebox{\linewidth}{!}{
    \scalebox{.50}{
        \begin{tabular}{lrr}
        \hline
        \ac{mllm} & ANet-QA  &  NExT-QA \\
        \hline
        No pseudo-labels & 53.5 & 62.4 \\
        LLaVA-NeXT 7B    & 53.9 & 62.8 \\
        Idefics2 8B      & 53.8 & 63.2 \\
        Qwen2 VL 7B      & 54.2 & 63.6 \\
        \hline
        \end{tabular}
        }
    }
    \caption{Performance of LLaVA-NeXT-Video 7B on ActivityNet (ANet) and NExT QA with different spatial pseudo-labels}
    \label{tab:appendix:1}
\end{table}

\begin{table}[ht]
    \centering
    \resizebox{\linewidth}{!}{
    \begin{tabular}{lrr}
    \hline
    \ac{mllm} & EgoSchema  &  LongVideoBench \\
    \hline
    Uniform 4 frames & 45.8 & 45.3 \\
    $16\rightarrow4$ & 47.8 & 46.0 \\
    $32\rightarrow4$ & 48.2 & 48.9 \\
    $128\rightarrow4$ & 49.0 & 49.5 \\
    \hline
    \end{tabular}
    }
    \caption{Performance of selecting different number of frames on EgoSchema and LongVideoBench with LLaVA-NeXT-Video 34B as downstream video-LLM.}
    \label{tab:appendix:2}
\end{table}

\vspace{-18pt}
\section{Additional Results}
\paragraph{Pseudo Label Geneation with different \ac{mllm}} Qwen2-VL serves as the prompting \ac{mllm} for spatial pseudo-labels generation as detailed in Section~\ref{sec:method.3}. We investigate the influence of alternative \ac{mllms} on video QA performance. Table~\ref{tab:appendix:1} compares the performance of LLaVA-Next-Video 7B on ActivityNet and NExt-QA using frames selected based on spatial pseudo-labels generated by different prompting \ac{mllms}. A stronger \ac{mllm} produces higher-quality pseudo-labels.

\vspace{-12pt}
\paragraph{Number of frames before selection} Existing frame selection methods~\citep{yu2024self, ranasinghe2024understanding} typically sample $16\sim32$ frames from a video and then perform frame selection on these frames. In contrast, our method samples a significantly larger list of 128 frames prior to the frame selection process. We posit that a larger number of frames is essential for long video QA. To evaluate this, we assessed the video QA performance of LLaVA-NeXT-Video 34B taking 4 frames selected from different number of frames. Table~\ref{tab:appendix:2} summarizes the results on the long-video QA benchmarks EgoSchema and LongVideoBench. The improvement on QA performance from $16\rightarrow4$ to $128\rightarrow4$ is significant, showing the necessity of have a large frame selection candidate pool.

\section{More Visualization Results} \label{supp:sec:viz}
\begin{figure*}[ht]
    \centering
    \includegraphics[width=\linewidth]{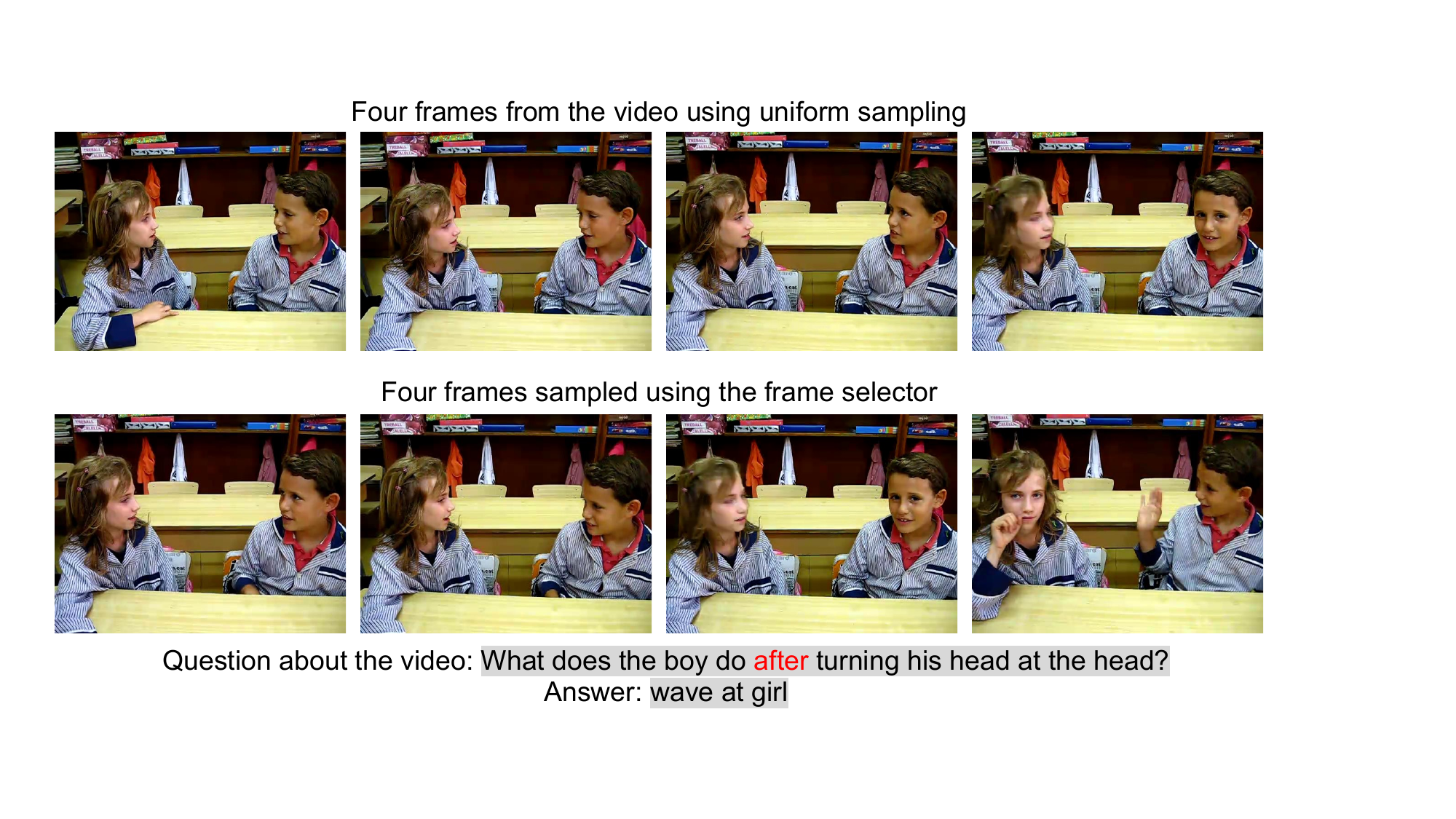}
    \caption{One visualization example of the frame selection results.}
    \label{fig:visual1}
\end{figure*}

\begin{figure*}[ht]
    \centering
    \includegraphics[width=\linewidth]{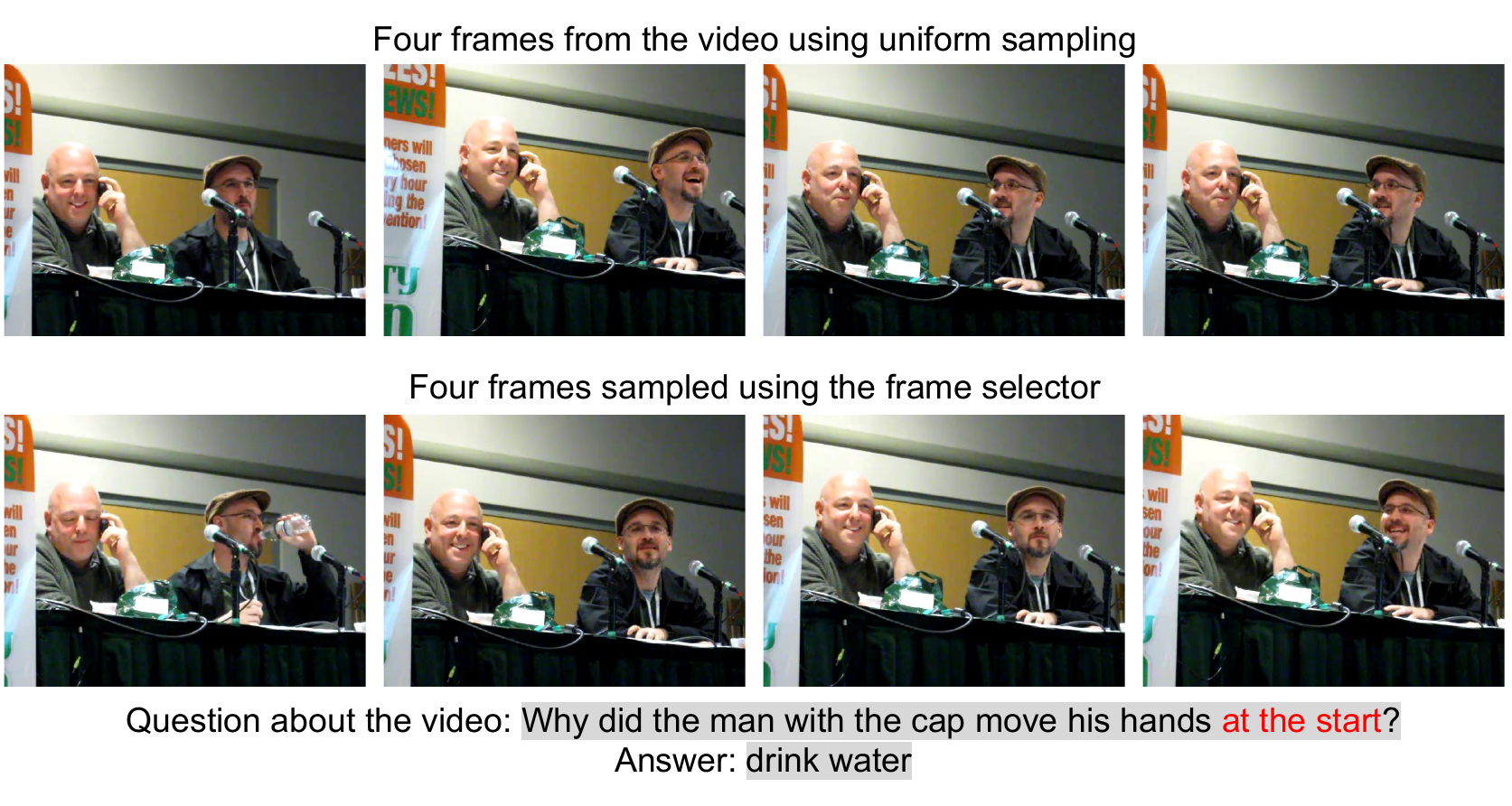}
    \caption{One visualization example of the frame selection results.}
    \label{fig:visual2}
\end{figure*}

\begin{figure*}[ht]
    \centering
    \includegraphics[width=\linewidth]{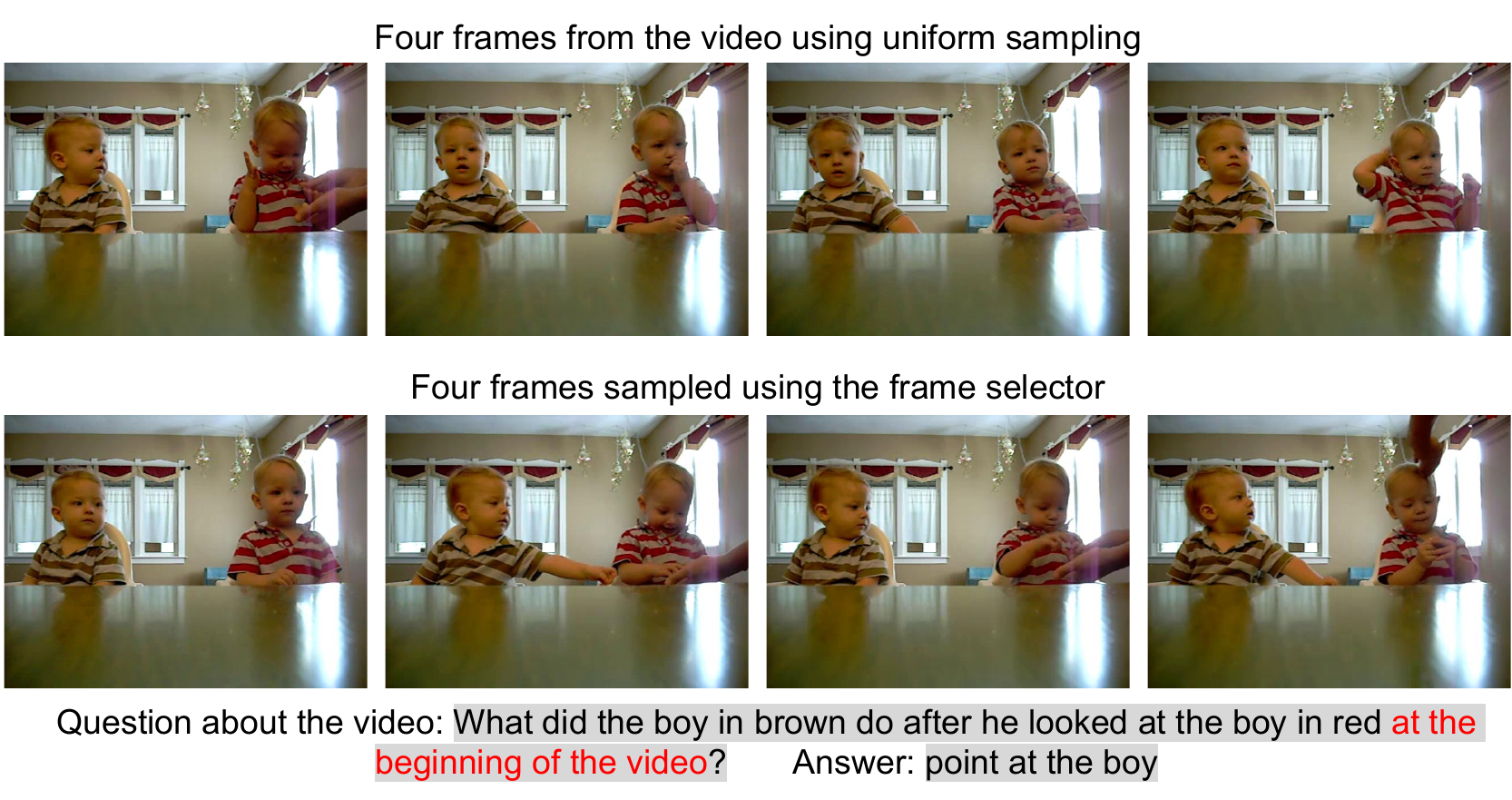}
    \caption{One visualization example of the frame selection results.}
    \label{fig:visual3}
\end{figure*}

\begin{figure*}[ht]
    \centering
    \includegraphics[width=\linewidth]{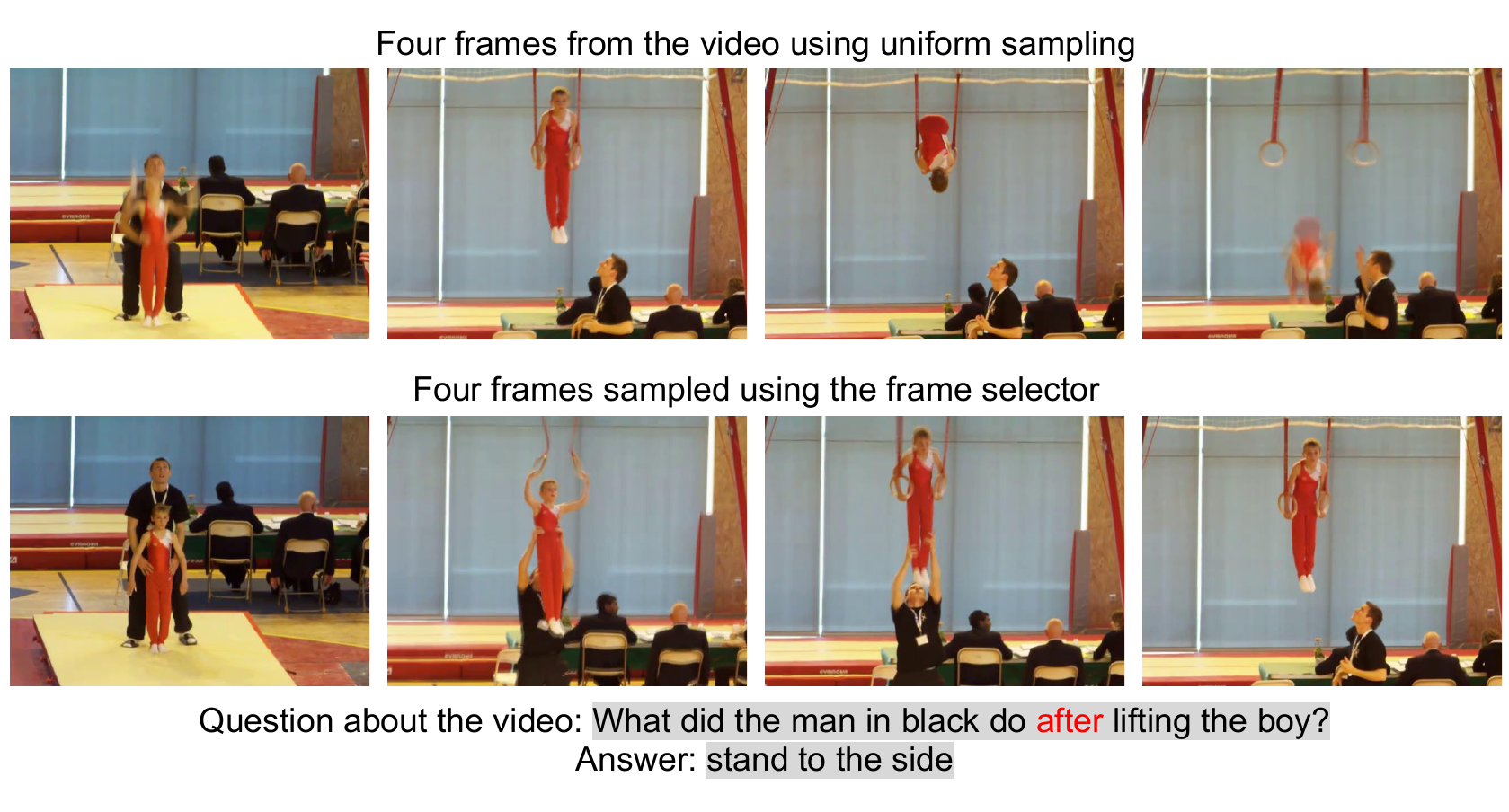}
    \caption{One visualization example of the frame selection results.}
    \label{fig:visual4}
\end{figure*}

Figure~\ref{fig:visual1} and Figure~\ref{fig:visual2} are the zoom-in for Figure~\ref{fig:visualazation}. Figure~\ref{fig:visual3} and Figure~\ref{fig:visual4} are additional visualization results.

\end{document}